\documentclass[letterpaper, 10 pt, journal, twoside]{IEEEtran}

\IEEEoverridecommandlockouts                              

\usepackage{amsmath,amssymb}
\usepackage{graphics} 
\usepackage{epsfig} 

\usepackage{lipsum}
\usepackage{mathtools}
\usepackage{cuted}
\usepackage{arydshln}
\usepackage{subcaption}
\usepackage{hyperref}

\usepackage{textcomp}
\usepackage{color}
\usepackage{url}
\usepackage{booktabs}
\usepackage{array}
\usepackage{tabularx}
\usepackage{xcolor}

\usepackage{bm}
\usepackage{siunitx}
\usepackage{graphicx}

\newcommand{\mat}[1]{\begin{bmatrix}#1\end{bmatrix}}
\newcommand{\revise}[0]{\textcolor{black}}
\newcommand{\revisesec}[0]{}

\definecolor{somegray}{rgb}{0.5, 0.5, 0.5}
\newcommand{\darkgrayed}[1]{\textcolor{somegray}{#1}}
\makeatletter
\newcommand*\titleheader[1]{\gdef\@titleheader{#1}}
\AtBeginDocument{%
  \let\st@red@title\@title
  \def\@title{%
    \vskip-2em
    \bgroup\normalfont\large\centering\@titleheader\par\egroup
    \vskip1.5em\st@red@title}
}
\makeatother

\titleheader{\darkgrayed{This paper has been accepted for publication at the\\
IEEE Robotics and Automation Letters, 2022.
\copyright IEEE}}
\title{ Nonlinear MPC for Quadrotor \\Fault-Tolerant Control}
\author{Fang Nan, Sihao Sun, Philipp Foehn, Davide Scaramuzza
\thanks{Manuscript received: September 09, 2021; Revised December, 19, 2021; accepted January 23, 2022.}
\thanks{
The authors are with the Robotics and Perception Group, Department of Informatics, University of Zurich, and Department of Neuroinformatics, University of Zurich and ETH Zurich, Switzerland (\protect\url{http://rpg.ifi.uzh.ch}).
This work was supported by the National Centre of Competence in Research (NCCR) Robotics, through the Swiss National Science Foundation (SNSF), and the European Union’s Horizon 2020 Research and Innovation Programme under grant agreement No. 871479 (AERIAL-CORE) and the European Research Council (ERC) under grant agreement No. 864042 (AGILEFLIGHT).
}
\thanks{Digital Object Identifier (DOI): see top of this page.}}
\begin{document}
\maketitle


\begin{abstract}
The mechanical simplicity, hover capabilities, and high agility of quadrotors lead to a fast adaption in the industry for inspection, exploration, and urban aerial mobility.
On the other hand, the unstable and underactuated dynamics of quadrotors render them highly susceptible to system faults, especially rotor failures.
In this work, we propose a fault-tolerant controller using nonlinear model predictive control (NMPC) to stabilize and control a quadrotor subjected to the complete failure of a single rotor. 
Differently from existing works, which either rely on linear assumptions or resort to cascaded structures neglecting input constraints in the outer-loop, our method leverages full nonlinear dynamics of the damaged quadrotor and considers the thrust constraint of each rotor.
Hence, this method could effectively perform upset recovery from extreme initial conditions. 
Extensive simulations and real-world experiments are conducted for validation, which demonstrates that the proposed NMPC method can effectively recover the damaged quadrotor even if the failure occurs during aggressive maneuvers, such as flipping and tracking agile trajectories.
\end{abstract}

\section*{Video}
Video of the experiments: \url{https://youtu.be/Cn_836XGEnU}

\section{Introduction}
\label{sec:introduction}
\subsection{Motivation}
Multirotor aerial vehicles are rapidly reshaping the industry of logistics, inspection, agriculture, and even flying cars~\cite{loianno2020special,air_taxi_2020}.
Until 2021, there have been more than 300,000 commercial drones registered in the United States alone, according to the Federal Aviation Administration~\cite{FAA2021UAS}, and the global drone market is projected to 63.6 billion USD by 2025~\cite{newswire_2019}.
However, flying accidents owing to system malfunctions, such as rotor failures, are major stumbling blocks to the development and the public acceptance of the drone industry.
For instance, two crashes in 2019 caused the suspension of the autonomous drone delivery service by the Swiss Post for a year in Switzerland, after more than 3,000 successful autonomous flights~\cite{ackermanSwiss2019}.

Fault-tolerant flight control (FTFC) is a promising technique to improve flying safety, which only requires algorithmic adaptations rather than mechanical changes (e.g., parachutes).
It is particularly valuable for quadrotors which are most susceptible to rotor failures, but are also the most popular type of all multirotors, due to their structural simplicity and efficiency.

\begin{figure}
    \centering
    \includegraphics[width=0.5\textwidth]{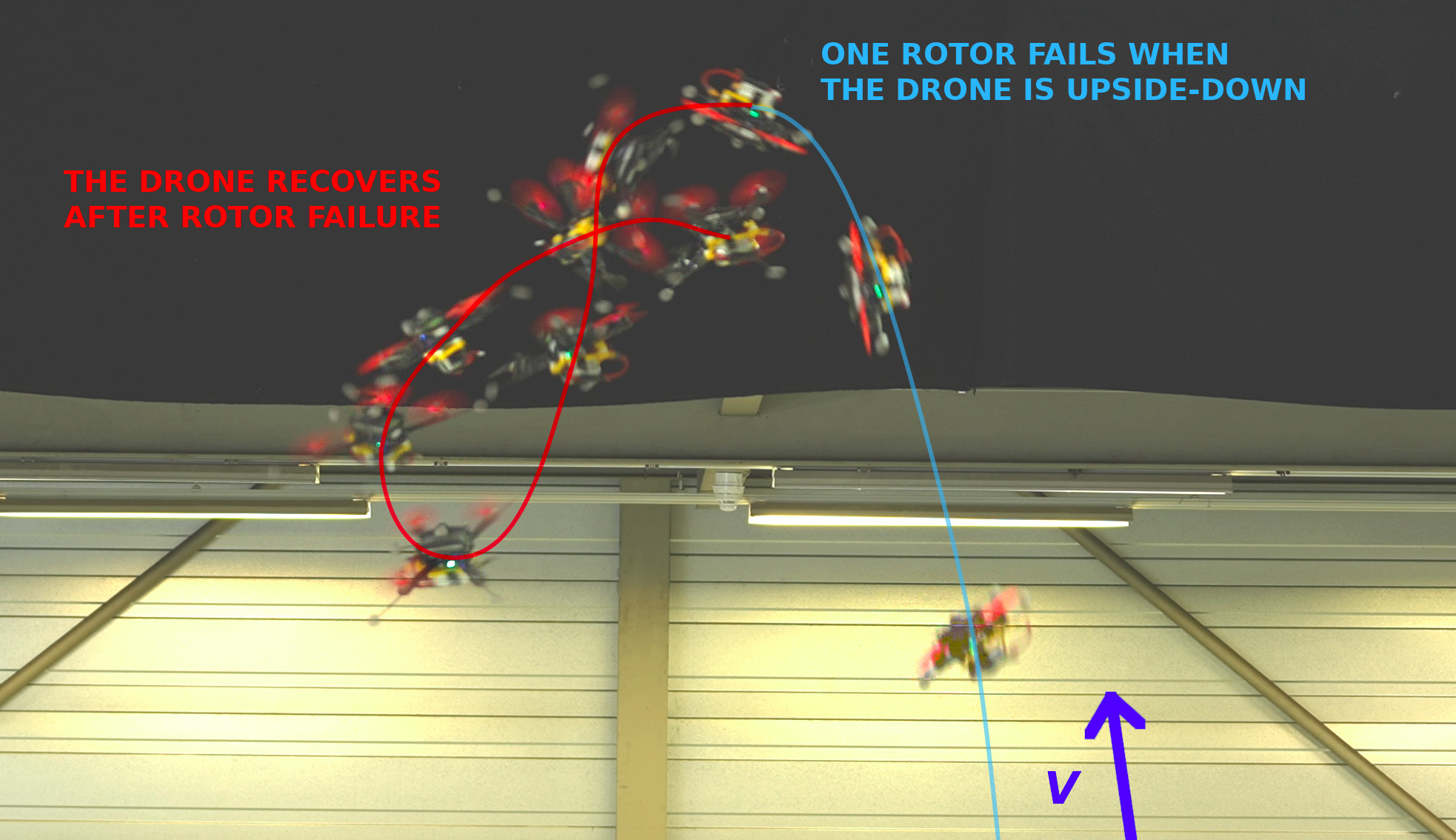}
    \caption{Our quadrotor recovering from a rotor failure occurred during a flipping maneuver. \textbf{One rotor completely failed} when the quadrotor was on the apex and upside-down, where our controller recovered the damaged quadrotor despite the rotor failure.}
    \label{fig:fliprecovery_snapshot}
\end{figure}

\subsection{Related Work}\label{sec:related_work}
FTFC is a long-standing challenge within the robotics and aerospace industries.
Initial research in this field focused on partial effectiveness loss of rotors, such as blade damage, where a quadrotor is mainly regarded as a research platform for testing novel robust/adaptive control methods.
These methods include but are not limited to sliding mode control~\cite{wang2019quadrotor,Besnard2012}, $\mathcal{L}_1$ adaptive control~\cite{Xu2016}, and active disturbance rejection control~\cite{Guo2018}.
Extensive research on fault detection and diagnosis is also carried out and tested on quadrotor platforms (e.g., see~\cite{amoozgar2013experimental,freddi2010actuator}).

While partial motor failure can be effectively handled by these adaptive methods, failure of a complete rotor is a more challenging problem and has a higher practical value.
This is because switching off a severely damaged rotor in practice is more favorable as it prevents vibrations caused by rotor imbalance.
A seminal work to address this problem is proposed by~\cite{freddi_feedback_2011}, which reveals that relinquishing the stability in yaw direction is inevitable after the failure of a motor.
But even then, the altitude and position of a post-failure quadrotor remain controllable.
Based on this seminal work \cite{freddi_feedback_2011} a wide variety of methods have been proposed and tested, including linear controllers such as proportional-integral-derivative (PID) controlller~\cite{lippiello_emergency_2014-1}, linear quadratic regulator (LQR)~\cite{mueller_stability_2014}, and linear parameter varying (LPV)~\cite{stephan_linear_2018} methods that rely on a relaxed hovering equilibrium to perform linearization~\cite{mueller2016relaxed}. 
\revise{The Sequential Linear Quadratic (SLQ) control \cite{de_crousaz_unified_2015}, on the other hand, performs the linearization around a predicted trajectory instead of a single equilibrium.}
\revise{Nonlinear methods that do not require the linearization} have also been proposed, including robust feedback linearization~\cite{lanzon_flight_2014}, nonlinear dynamic inversion (NDI)~\cite{sun2021autonomous}, backstepping~\cite{lippiello_emergency_2014}, and incremental nonlinear dynamic inversion (INDI)~\cite{sun_incremental_2020,sun_high-speed_2018}.
Recently, vision-based approaches proposed by~\cite{sun2021autonomous} have also jointly addressed the control and state estimation problem using only onboard sensors.

\revise{Most of} these methods, however, rely on different assumptions for controller design, such as small-angle assumptions for model linearization or the principle of time-scale separations for separately designing position and attitude controllers.
Moreover, \revise{all the aforementioned controllers} do not consider actuator limits, which are crucial for flight performance, especially after rotor failures.
For this reason, the nonlinear model predictive controller \revise{(NMPC)}, which exploits the full dynamics of a quadrotor, is a perfect candidate to tackle the fault-tolerant control problem. 

NMPC is a control method that uses a nonlinear model to predict the system dynamics from a current time over a receding horizon and optimizes the behavior of the system based on certain performance measurements.
It was shown in previous research that NMPC on quadrotors has a great advantage in agile trajectory tracking and dealing with constraints \cite{Foehn_2021, bangura_real-time_2014, kamel_fast_2015}. 
Due to the ability of the NMPC to fully exploit the input space without constraint violation, it was tested as the FTFC to stabilize a hexacopter with a rotor failure in~\cite{kamel_fast_2015}, although the loss of a single rotor is not a challenging situation for hexacopter control.
A more difficult case was tested, in which three rotors failed on a hexacopter~\cite{aoki_nonlinear_2018}.
It was shown with simulations that the nonlinear NMPC controls the horizontal position and maintains the altitude of the drone.
The authors of \cite{jung_fault_2021} also proposed a method using MPC as the inner loop controller for a quadrotor subject to rotor failures.
However, in \cite{aoki_nonlinear_2018} and \cite{jung_fault_2021}, the method is only validated in the simulation, and only tested for FTFC starting from near-hovering conditions. 

NMPC has several disadvantages that make it difficult to be implemented in the real world \cite{darby_mpc_2012, eren_model_2017}.
To obtain good predictions it is necessary to accurately model the system dynamics, which is particularly challenging for aerial robots whose dynamics are severely influenced by aerodynamic effects.
The complexity of the optimization problem also makes NMPC difficult to satisfy the requirements for real-time control.
As a result, \revise{to the best of our knowledge, }NMPC has never been implemented in real-world experiments to control a quadrotor in the case of a complete failure of a rotor.



\subsection{Contribution}
We present a nonlinear model predictive control (NMPC) framework for quadrotors that is tolerant to the complete failure of a single rotor. \revise{Using a real-time iteration scheme, the NMPC can be solved onboard at an adequate speed that satisfies the requirements for control.}
The proposed controller considers the full nonlinear dynamics of quadrotors, without resorting to a cascaded structure or any assumptions for linearization.
The NMPC uses single rotor thrusts as inputs and considers their correct limits, which allows exploiting the full potential of any intact and damaged quadrotor without violating the real dynamics. \revise{An INDI low-level controller is applied to regulate the original NMPC thrust command to compensate for model uncertainties and disturbances on the rotational dynamics.}


Experimental validations are performed in both simulation and real-world environments.
We demonstrate the recovery performance of the proposed controller from several extreme conditions, such as flipping maneuvers and agile flights.
Experiments also validate the ability of our controller in tracking agile trajectories despite the complete failure of a single rotor.
Such recovery performance can significantly expand the flight envelope of damaged quadrotors, enabling controlled descent and landing maneuvers and ultimately allowing even safe flight under rotor failures.


\section{Methodology}\label{sec:methodologies}
This section will introduce the control algorithm, and the model used for controller design. 
We use bold lowercase letters for vectors (e.g., $\bm{v}$) and bold uppercase letters for matrices (e.g., $\bm{M}$); otherwise they are scalars.
Two right-handed coordinate systems are used to represent the body frame $\{\bm{x}_B,~\bm{y}_B,~\bm{z}_B\}$ and the inertial frame $\{\bm{x}_I,~\bm{y}_I,~\bm{z}_I\}$. 
The body frame is defined as is shown in Fig.~\ref{fig:quad_model}. 
The inertial frame is defined with $\boldsymbol{z}_I$ pointing upwards opposite to gravity.

\subsection{Quadrotor Modeling}
\label{sec:quad_modeling}
\begin{figure}[h]
  \centering
  \includegraphics[width=0.7\columnwidth]{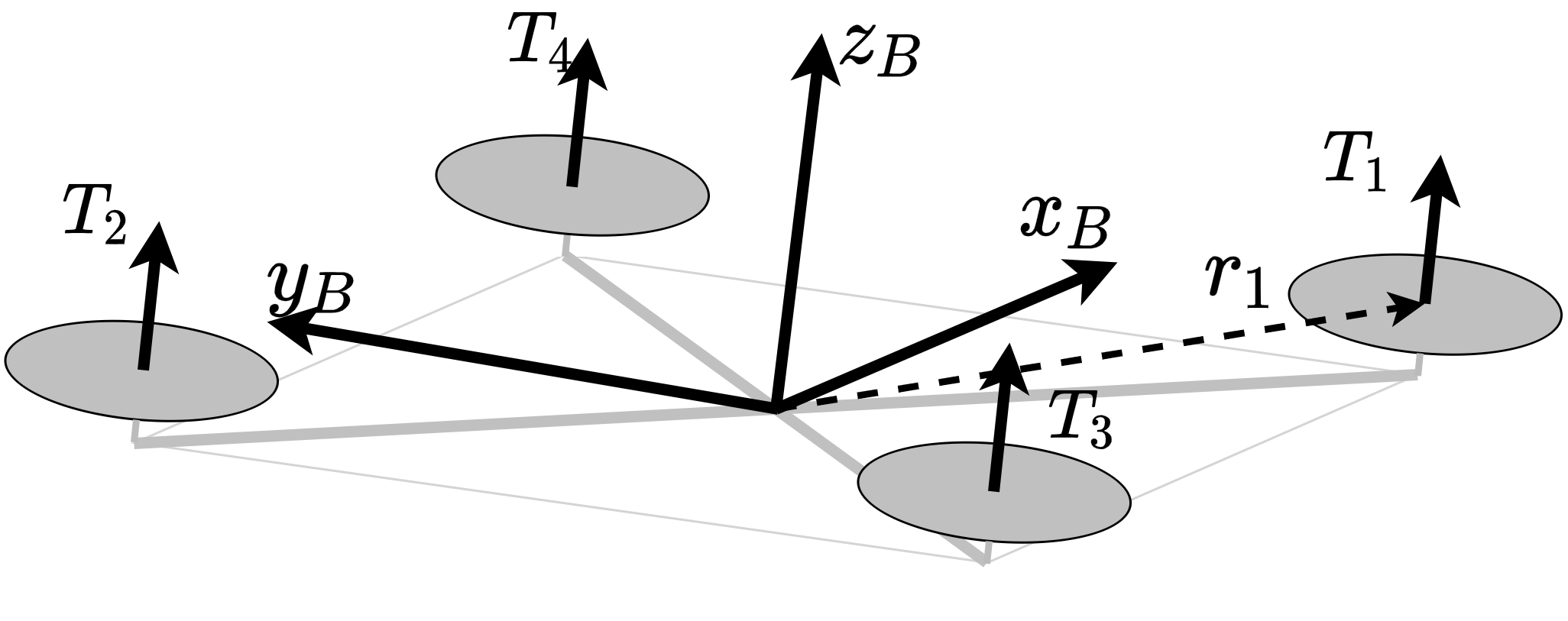}
  \caption{Definition of quadrotor body frame and rotor indices.}
  \label{fig:quad_model}
\end{figure}
To model the dynamics of a quadrotor, we define the quadrotor body frame as shown in Fig.~\ref{fig:quad_model}.
We use $T_1$ to $T_4$ to represent the thrusts generated by the four rotors, with the indices shown in Fig.~\ref{fig:quad_model}, where  $\bm{r}_1$ to $\bm{r}_4$ denote the position vectors of these rotors.
Using quaternion representation of the quadrotor orientation, we could write the dynamics of the quadrotor as
\begin{gather}
\begin{aligned}
  \dot{\bm{p}} & = \bm{v}, &
  \dot{\bm{q}} & = \frac{1}{2} \bm{q} \circ \left[\begin{matrix} 0 \\ \bm{\omega}\end{matrix}\right], \\
  \dot{\bm{v}} &= \bm{q} \odot \begin{bmatrix} 0 \\ 0 \\ T \end{bmatrix} - \bm{g}, &
  \dot{\bm{\omega}} & = \bm{I}_v^{-1} \left(\bm{\tau} - \bm{\omega} \times (\bm{I}_v \bm{\omega})\revise{+\bm{\tau}_\mathrm{ext}}\right),
  \label{eqn:quad_model}
\end{aligned}
\end{gather}
where $\bm{p}$ and $\bm{v}$
are the position and velocity of the center of gravity expressed in the inertial frame.
$\bm{q} = \begin{bmatrix} q_w & q_x & q_y & q_z \end{bmatrix}^T$ is the quaternion representation of the quadrotor orientation, and $\bm{\omega} = \left[ \begin{matrix} \omega_x & \omega_y & \omega_z \end{matrix} \right]^T$ is the angular velocity of the quadrotor expressed in the body frame.
$\bm{I}_v$ is the inertia matrix of the entire quadrotor.
The $\circ$ symbol represents quaternion multiplication, while $\odot$ represents the rotation of a vector \revise{using a quaternion.
The unmodeled external torques due to aerodynamic disturbances and model mismatch are denoted as $\bm{\tau}_\mathrm{ext}$.}
$T$ and $\bm{\tau}$ are collective thrust and torque \revise{generated by the actuators}, expressed by
\begin{equation}
    \left[\begin{array}{c}
         T  \\
         \bm{\tau} 
    \end{array}\right] = \bm{G}\bm{t},
    \label{eq:allocation_equation}
\end{equation}
where $\bm{t} = \left[\begin{matrix} T_1 & T_2 & T_3 & T_4 \end{matrix}\right]^T$ \revise{is the thrust vector}, and $\bm{G}$ is the control effectiveness matrix:
\begin{equation}
    \bm{G} = \left[\begin{array}{cccc}
         1 & 1& 1 & 1  \\
         r_{y,1}&r_{y,2}&r_{y,3}&r_{y,4}\\
         -r_{x,1}&-r_{x,2}&-r_{x,3}&-r_{x,4}\\
         -\kappa_t&-\kappa_t&\kappa_t&\kappa_t
    \end{array}\right],
\end{equation}
where $r_{x,i}$ and $r_{y,i}$ indicate the x and y component of $\boldsymbol{r}_i$ in the body frame, and $\kappa_t$ represents the torque coefficient, a factor between thrust and drag torque generated by a single rotor. 

The model in (\ref{eqn:quad_model})-(\ref{eq:allocation_equation}) takes the single rotor thrusts as input.
However, when performing real-time control, the motors cannot track the commanded thrust without delay.
Thus we also included motor dynamics into the model.
We used the first-order model on the rotor thrust to approximate the motor dynamics, yielding 
\begin{equation} \label{eqn:motor_model}
  \dot{T}_i = \frac{1}{\sigma} (u_{i} - T_i), \qquad i = 1, 2, 3, 4,
\end{equation}
where $\sigma$ is the time constant identified from data, and $u_i$ is the command for the $i$-th rotor.

\subsection{Fault-Tolerant Model Predictive Control}
\label{sec:mpc}


\revise{We define the state of the quadrotor $\bm{x} = [\bm{p}^T \  \bm{q}^T \ \bm{v}^T \ \bm{\omega}^T \ \bm{t}^T]^T$ and the input $\bm{u} = [u_1 \ u_2 \ u_3 \ u_4]^T$. Then the discrete dynamics of the quadrotor $\bm{x}_{k+1} = f(\bm{x}_k, \bm{u}_k)$ could be obtained by discretizing (\ref{eqn:quad_model}) and (\ref{eqn:motor_model}).}

NMPC optimizes the following cost function with constraints on system dynamics and input bounds to solve for the optimal control input sequence $\bm{u}_{k:k+N-1}$:
\begin{equation}
\begin{aligned}
  \min_{\bm{u}_{k:k+N-1}} & \bm{y}_N^T \bm{Q}_N \bm{y}_N +  \sum_{i=k}^{k+N-1} \bm{y}_i^T \bm{Q} \bm{y}_i + \bm{u}_i^T \bm{R} \bm{u}_i
  \\
  \revise{\text{s.t.}} & \revise{\quad \bm{x}_{i+1} = f(\bm{x}_i,\bm{u}_i) \quad i=k,k+1,\dots, k+N-1} \\
  & \revise{\quad \underline{\bm{u}} \leq \bm{u} \leq \bar{\bm{u}}. }
  \end{aligned}
\end{equation}
where $k$ is the current time step, $N$ is the number of sampled time steps in the prediction horizon, and $\bm{x}_{k:k+N}$ is the predicted state trajectory. \revise{$\bm{Q}$, $\bm{Q}_N$ and $\bm{R}$ are positive-definite weight matrices. $\underline{\bm{u}} = \bm{0}_{4\times1}$ is the lower bound of the thrust command, while $\bar{\bm{u}}$ is the upper bound of the thrust command.
During normal flight $\bar{\bm{u}} = T_{max}\bm{1}_{4\times1}$, and if the $i$-th rotor fails, we set the $i$-th element of $\bar{\bm{u}}$ to $\bar{\bm{u}}_i = \underline{\bm{u}}_i = 0$.
Hence NMPC is aware of the fact that the $i$-th motor is unable to generate thrust.}

The cost function consists of two components: a running cost that accumulates at every time step as a function of $\bm{x}$ and $\bm{u}$, and a terminal cost that only depends on the terminal state.

For a quadrotor with a failed rotor, it is impossible to fully control the attitude. In order to define the cost on attitude, we first compute the attitude error $\bm{q}_w = \bm{q}_{ref} \circ \bm{q}^{-1}$ from the reference ($\bm{q}_{ref}$) and actual ($\bm{q}$) attitude.
then we split the attitude error into $z$ and $xy$ rotation according to $\bm{q}_e = \bm{q}_z \circ \bm{q}_{xy}$ where $\bm{q}_z = \mat{ q_{z,w} & 0 & 0 & q_{z,z}}^\intercal$ and $\bm{q}_{xy} = \mat{q_{xy,w} & q_{xy,x} & q_{xy,y} & 0}^\intercal$.
We could then form the cost vector $\bm{y}_i$ and the block-diagonal weight matrix $\bm{Q}$ for the running costs as
\begin{equation}
\bm{y}_i = \mat{
\bm{p} - \bm{p}_{ref} \\
q_{xy,x}^2 + q_{xy,y}^2 \\
q_{z,z} \\
\bm{v} - \bm{v}_{ref} \\
\bm{\omega} - \bm{\omega}_{ref} \\
\bm{t} - \bm{t}_{ref} \\
\bm{u} - \bm{u}_{ref}
},
\end{equation}
\begin{equation}
\bm{Q} = \mathrm{diag} \left( \mat{
\bm{Q}_p &
Q_{xy} & Q_{z} &
\bm{Q}_v &
\bm{Q}_\omega &
\bm{Q}_t &
\bm{Q}_u &
} \right),
\label{eq:weights}
\end{equation}
with all values taken at their time-step $i$.
Additionally, the terminal cost $\bm{y}_N^T \bm{Q}_N \bm{y}_N$ is structured the same way, but without the input $\bm{u}$.
%
%
%
The reference values in the cost functions are either hover references used for all time points in the prediction horizon, or sampled points from trajectories.


We set $Q_z$ in (\ref{eq:weights}) to zero in the rotor-failure condition, \revise{and only keep the cost on $\bm{q}_{xy}$. Thus we enforce the quadrotor to align its thrust direction with the reference while discarding the control of yaw motion.}

\subsection{Incremental Nonlinear Dynamic Inversion (INDI)}
\label{sec:indi}
For quadrotors fault-tolerant control, the aerodynamic effects arising from the high body rates cannot be accurately described by a model that is simple enough to be inserted into the NMPC.
In addition, the center of gravity position and other mechanical, electrical, or environmental effects can all result in further model mismatches.

{\revise{In order to mitigate these effects, similar to previous research~\cite{sun2021comparative}, INDI has been used to refine the thrust command from NMPC to be adaptive against model uncertainties. Specifically, INDI approximates $\bm{\tau}_\mathrm{ext}$ in (\ref{eqn:quad_model}) by instantaneous sensor measurements instead of finding an accurate model that can be difficult to obtain.}
There are different derivations of INDI for quadrotors (see, e.g., \cite{smeur2016adaptive,tal2020accurate}). Here we adopt the version from ~\cite{tal2020accurate}. According to the last equation of ($\ref{eqn:quad_model}$), we can estimate the external torques as
\begin{equation}
    \bm{\tau}_\mathrm{ext} = \bm{I}_v\dot{\bm{\omega}}_f - \bm{\tau}_f + \bm{\omega}_f \times \bm{I}_v\bm{\omega}_f
    \label{eq:tau_ext_INDI}
\end{equation}
where the subscript $f$ indicates that the variable is measured and low-pass filtered for noise reduction. $\bm{\tau}_f$ is the estimated torque obtained from (\ref{eq:allocation_equation}), with the force of each rotor $\bm{t}$ obtained from the low-pass filtered rotor speed measurements. Note that all variables have a similar amount of delay as they are filtered with the same LPF. We also assume that the bandwidth of $\bm{\tau}_\mathrm{ext}$ is significantly lower than the bandwidth of the LPF. Thus the delay of $\bm{\tau}_\mathrm{ext}$ is negligible. Then we substitute (\ref{eq:tau_ext_INDI}) into the last equation of (\ref{eqn:quad_model}), yielding
\begin{equation}
\begin{aligned}
    \bm{I}_v\dot{\bm{\omega}} =&~\bm{I}_v\dot{\bm{\omega}}_f + \bm{\tau} - \bm{\tau}_f + \left(\bm{\omega}_f\times\bm{I}_v\bm{\omega}_f - \bm{\omega}\times\bm{I}_v\bm{\omega}\right) \\
    \approx&~\bm{I}_v\dot{\bm{\omega}}_f + \bm{\tau} - \bm{\tau}_f
\end{aligned}
\label{eq:INDI1}
\end{equation}
The approximation in (\ref{eq:INDI1}) is reasonable since the actuator dynamic is significantly faster than the quadrotor body rate change. 

With the thrust command $\bm{u}$ from NMPC, we can obtain the desired collective thrust $T_d$ and angular acceleration $\bm{\alpha}_d$ from (\ref{eqn:quad_model}) and (\ref{eq:allocation_equation}), yielding 
\begin{equation}
    \left[\begin{array}{c}
         T_d  \\
         \boldsymbol{I}_v\bm{\alpha}_d + \bm{\omega}\times\bm{I}_v\bm{\omega} 
    \end{array}\right] = \bm{G}\bm{u}
    \label{eq:INDI0}
\end{equation}

Replacing the left-hand side of (\ref{eq:INDI1}) by the desired angular acceleration $\bm{\alpha}_d$ obtained from (\ref{eq:INDI0}), we can calculate the desired torque:}
\begin{equation}
    \boldsymbol{\tau}_d = \bm{\tau}_f + \bm{I}_v \left(\bm{\alpha}_d - \dot{\boldsymbol{\omega}}_f\right),
    \label{eq:INDI2}
\end{equation}
Finally, the thrust command of each rotor can be obtained by 
\begin{equation}
    \bm{u}_{indi} = \bm{\hat{G}}^{+}\left[\begin{array}{c}
         T_d  \\
         \bm{\tau}_d 
    \end{array}\right] 
    \label{eqn:indi_formular}
\end{equation}
where $\bm{\hat{G}}$ is the reduced control effectiveness matrix by setting the $i$-th column of $\boldsymbol{G}$ to zero.
As such, the final thrust command $\bm{u}_{indi}$ of the damaged ($i$-th) rotor is always zero.
Since $\bm{\hat{G}}$ is not full rank, we use Moore-Penrose inversion in (\ref{eqn:indi_formular}).

\section{Simulations}
\label{sec:simulations}

To validate the effectiveness of the proposed fault-tolerant controller, we first conduct experiments in a simulation environment. \revise{We also compare the proposed NMPC against a benchmark method~\cite{sun_incremental_2020} in the same simulation environment. The benchmark is a cascaded nonlinear controller that has been validated in real-world experiments withstanding strong external wind disturbances.} Afterward, real-world experiments will be performed and described in Sec.~\ref{sec:experiments}.

\begin{table}[h]
    \centering
    \setlength{\tabcolsep}{3pt}
    \revise{\begin{tabular}{c c c}
        \toprule
        Method & Avg. iter time (\si{\ms}) & Max. iter time (\si{\ms}) \\
        \midrule
        Benchmark & 1.122e-2 & 0.5459 \\
        NMPC & 6.431 & 25.75 \\
        \bottomrule
    \end{tabular}}
    \caption{\revise{Average and maximum iteration computing of proposed NMPC controller and the benchmark controller. Measured during simulations on a Jetson TX2.}}
    \label{tab:iter_time_comparison}
\end{table}


\begin{table}[h]
    \centering
    \setlength{\tabcolsep}{3pt}
    \revise{
    \begin{tabular}{c c | c c}
        \toprule
        Weight & Value & Weight & Value \\
        \midrule
        $\bm{Q}_{p,xy}$ & 80 & $\bm{Q}_{p,z}$ & 800 \\
        $\bm{Q}_{xy}$ & 60 & $\bm{Q}_v$ & 1 \\
        $\bm{Q}_{\omega, xy}$ & 0.5 & $\bm{Q}_{\omega, z}$ & 0.1 \\
        $\bm{Q}_t$ & 3.0 & $\bm{Q}_u$ & 1.0 \\
        \bottomrule
    \end{tabular}}
    \caption{\revise{Weight selection of the NMPC cost function}}
    \label{tab:mpc_weight}
\end{table}


\subsection{Implementation Details}

\revise{For both simulations and real-world experiments, the control algorithm is implemented in Agilicious \cite{robotics_and_perception_group_agilicious_nodate}, a software and hardware co-designed platform for agile autonomous quadrotor flight.
The NMPC controller is solved by a sequential quadratic programming (SQP) algorithm executed in a real-time iteration scheme~\cite{diehl2006fast}.
Specifically, we implement the algorithm using ACADO \cite{Houska2011a} with qpOASES~\cite{ferreau2014qpoases} as the solver.}
\revise{In both the simulations and the real-world experiments, we set the weight of the NMPC cost function to be the values in Table \ref{tab:mpc_weight}.}
\revise{A prediction horizon of \SI{1}{\second} is used, and discretized into 20 steps.
The position tracking error is clamped to prevent divergence of the NMPC solver.}

\revise{A mode switch mechanism is added to the software to simulate the rotor failure.
Once the switch is triggered externally, the controller command for one of the motors is clamped to zero immediately.
Since there are many effective and rapid quadrotor failure detection methods (see, e.g., \cite{freddi2010actuator,amoozgar2013experimental}), the fault-detection problem is out of the scope of this paper.
Instead, the index of the failed rotor is directly sent to the flight controller when the failure is triggered through a mock fault detector.}

\revise{In order to evaluate the computation time of the proposed and the benchmark method, the simulations have been performed on the NVIDIA Jetson TX2, the same onboard computing unit we use in the real-world experiments. The computational time for the controller is recorded and listed in Table \ref{tab:iter_time_comparison}. It is clear from the data that the NMPC has a much longer iteration time than the nonlinear controller. However, on average the frequency of the NMPC reaches more than \SI{150}{\hertz}, which we found enough for the control of quadrotor in the tests. The maximum iteration time of the NMPC reaches \SI{25}{\ms}, but it only appears at the first iteration after the rotor failure is triggered. As the real-time iteration scheme is used, the internal states of the solver have to be reinitialized to ensure no violation of the updated constraint. Thus this particular iteration takes a longer time than average.}

\subsection{Recovery from Randomized Orientations}

\begin{figure}[h]
    \centering
    \begin{subfigure}[b]{0.9\columnwidth}
    \includegraphics[width=\textwidth]{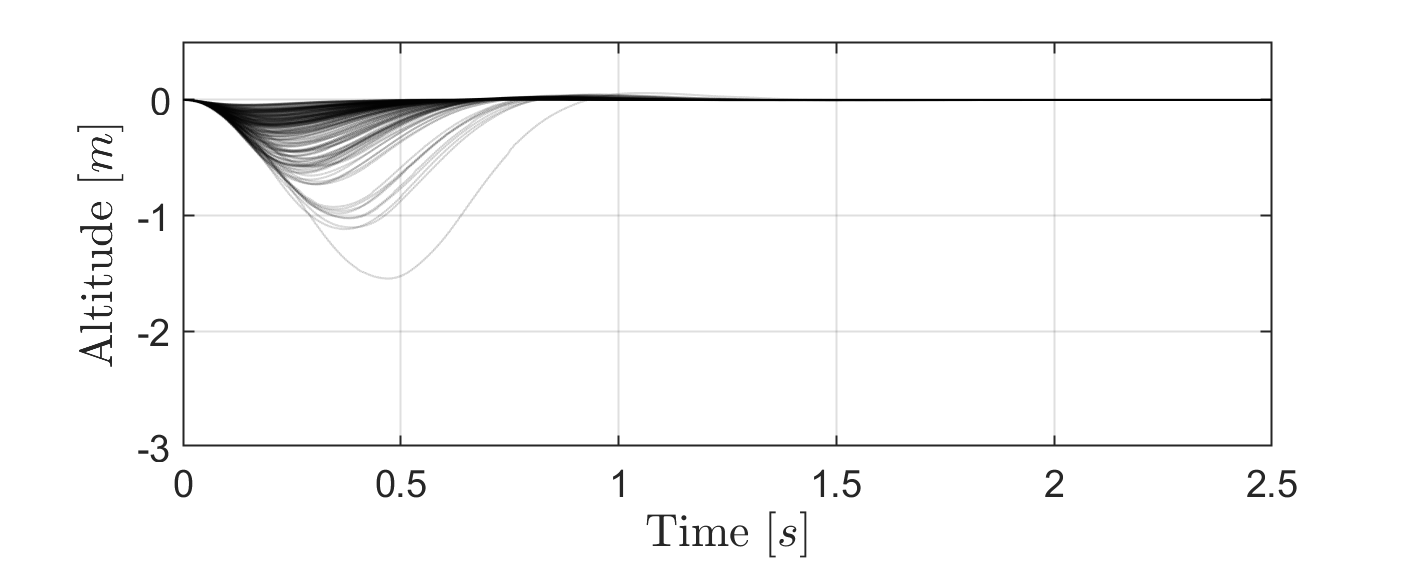}
    \caption{NMPC}
    \label{fig:randrec_mpc_z}
    \end{subfigure}
    \\
    \begin{subfigure}[b]{0.9\columnwidth}
    \includegraphics[width=\textwidth]{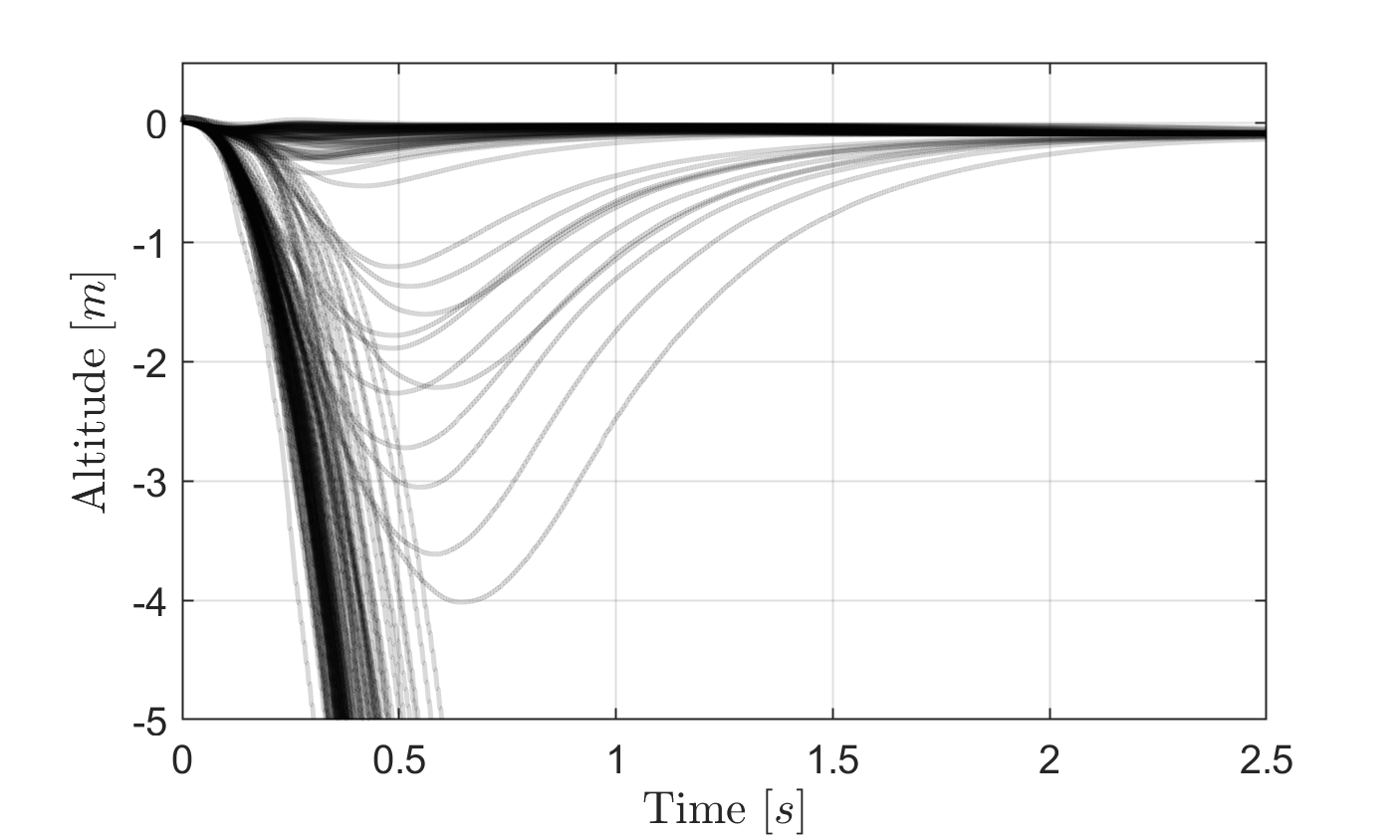}
    \caption{Benchmark}
    \label{fig:randrec_bcmk_z}
    \end{subfigure}
    \caption{\revise{Time history of the relative altitude with respect to the initial point in the randomized recovery tests using (a) proposed NMPC, and (b) benchmark controller.}}
    \label{fig:randrecover_z_mpcvsbcmk}
\end{figure}

\begin{figure}[h]
    \centering
    \includegraphics[width=\columnwidth]{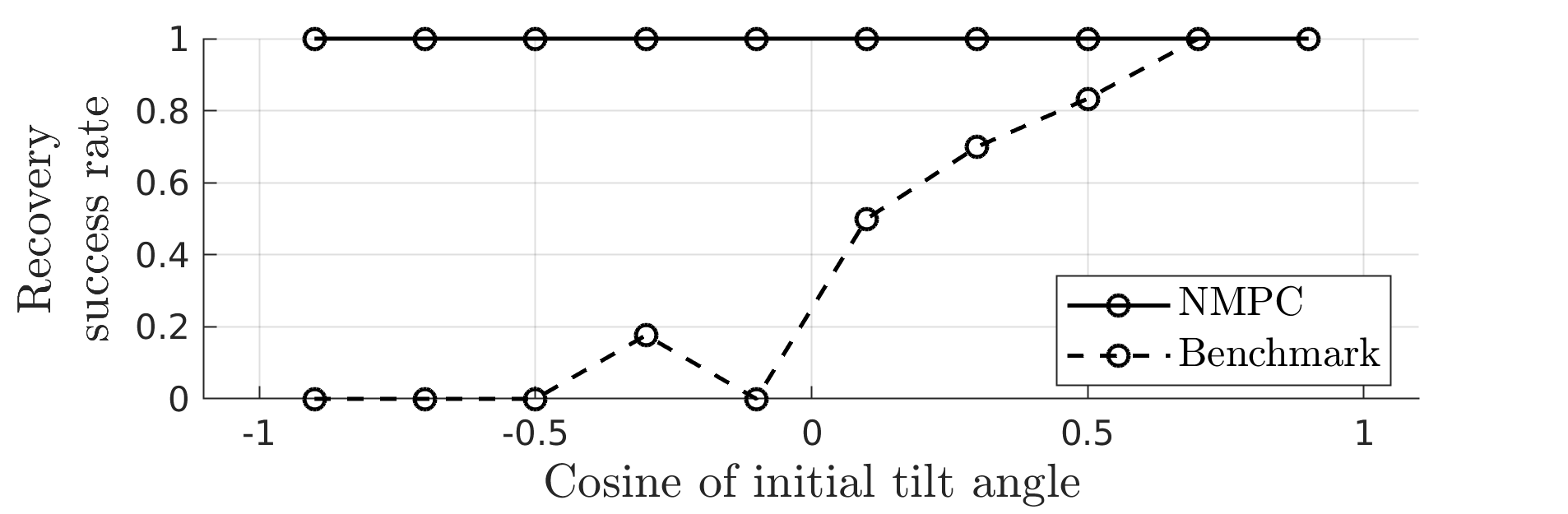}
    \caption{\revise{The success rate of recovery using the proposed NMPC and the benchmark method. Each point shows the success rate of recovery when the cosine of the initial tile angle falls in the corresponding interval.}}
    \label{fig:randrec_success_rate}
\end{figure}

\begin{figure}[h]
    \centering
    \begin{subfigure}[b]{0.5\columnwidth}
    \includegraphics[width = \textwidth]{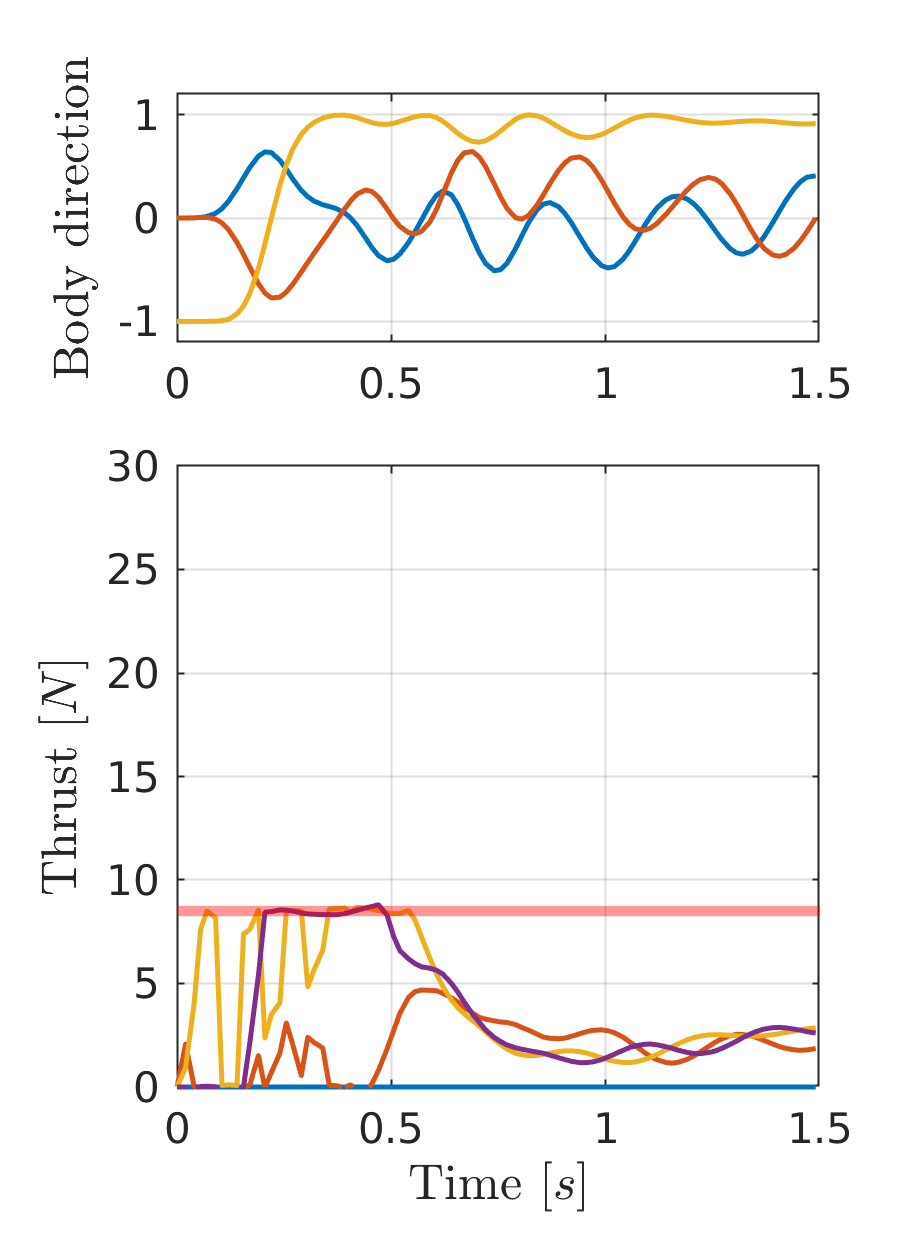}
    \caption{NMPC}
    \label{fig:mpc_nosaturation}
    \end{subfigure}
    \begin{subfigure}[b]{0.48\columnwidth}
    \includegraphics[width=\textwidth]{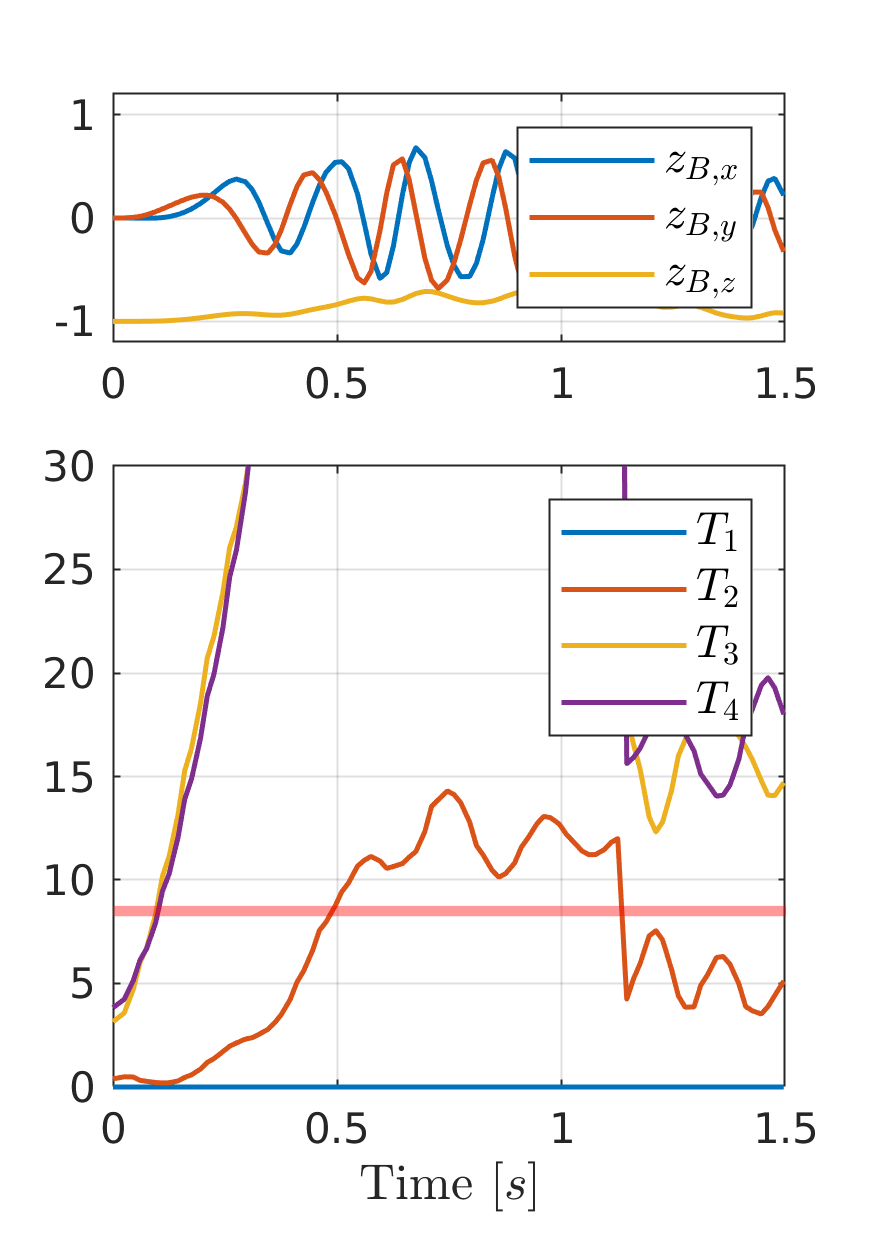}
    \caption{Benchmark}
    \label{fig:bcmk_saturation}
    \end{subfigure}
    \caption{\revisesec{Quadrotor attitude and thrust commands during a recovery test in simulation. Initial attitude is \SI{180}{\degree} flipped. \textbf{(a)} controlled by NMPC, \textbf{(b)} controlled by the benchmark controller. The bold red line in the thrust plots shows the actuator limit. The attitude of the quadrotor is represented by the $x$, $y$, and $z$ components (in the inertial frame) of $z_B$, a unit vector in the thrust direction.}}
    \label{fig:mpcvsbenchmark_saturation}
\end{figure}

\revise{Extensive randomized simulations are performed to compare the ability to recover a damaged quadrotor from different initial orientations.}
\revise{In the simulations, the quadrotor is initialized at random orientations with a rotor failure and commanded to hover at the same position. The initial orientations of the quadrotor are sampled from a  uniform distribution~\cite{ARVO1992117}, and the velocity and angular rates are initialized with zeros.}

\revise{The recovery simulation is performed 200 times with both the proposed NMPC and the benchmark method. The altitude profiles in these flights are shown in \ref{fig:randrecover_z_mpcvsbcmk}.
In all 200 randomized tests, the quadrotor is able to recover from the initial orientation when controlled by the NMPC. With the benchmark method, however, it only succeeded in 85 of the tests.
The recovery time and the loss of height in the recovery process are also significantly lower using the proposed NMPC.
We sorted the 200 tested initial orientations based on the cosine of the tilt angle, and computed the success rate of the recoveries in different intervals of initial tile angles. The success rate is plotted in Fig.~\ref{fig:randrec_success_rate}.
The data shows that the benchmark method is only able to recover the quadrotor if the initial orientation is close to the hovering orientation, while the success rate is almost zero if the quadrotor's initial direction is more than \SI{90}{\degree} tilted.
}
\revisesec{Figure \ref{fig:mpcvsbenchmark_saturation} shows the attitude of the quadrotor when trying to recover it from an upside-down direction with either controller. It is observed that the benchmark controller, which does not take into account the actuator limits, fails to recover the quadrotor after experiencing actuator saturation. NMPC, however, recovers the quadrotor without violating the actuator constraints.}

\section{Real-World Experiments}\label{sec:experiments}

\subsection{Experiment Setup}\label{sec:experiment_setup}
\begin{figure}[t]
  \centering
  \includegraphics[width=0.9\columnwidth]{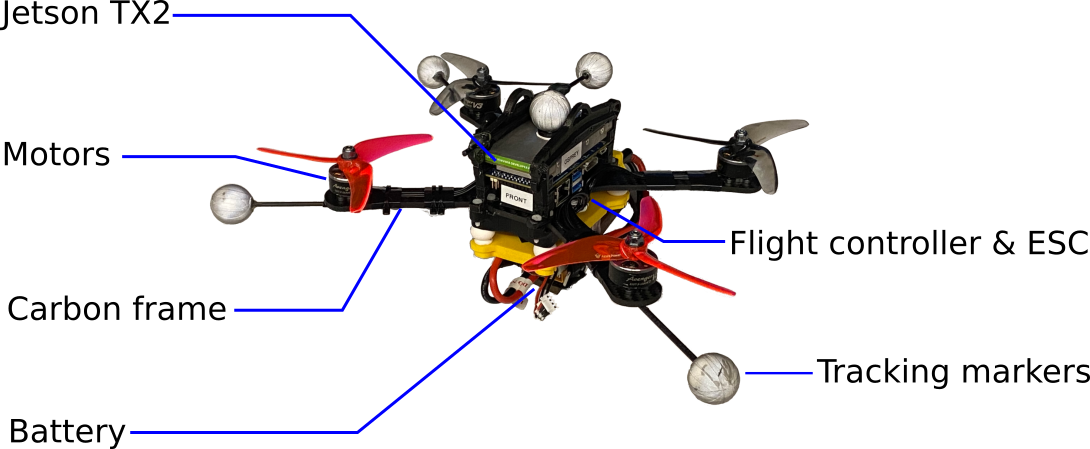}
  \caption{Photo of the tested quadrotor platform.}
  \label{fig:drone_photo}
\end{figure}

\begin{figure}[t]
  \centering
  \includegraphics[width=\columnwidth]{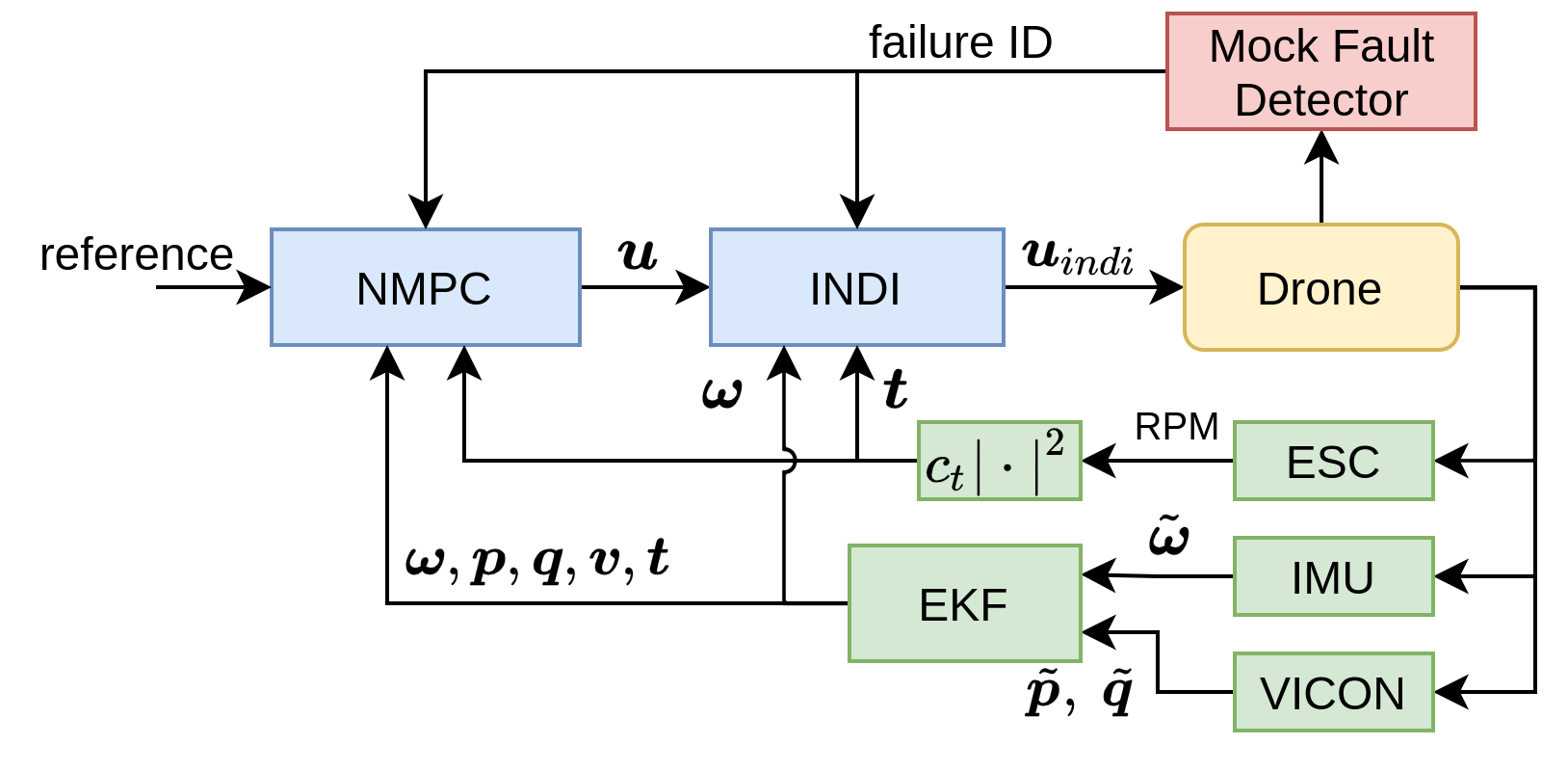}
  \caption{System diagram in real-world experiments. An NMPC block generates the initial thrust command of rotors, followed by an INDI block to modify these commands for robustification.
  The full states are obtained by an extended Kalman filter (EKF) fusing measurements from an inertia measurement unit (IMU) and a motion capture system (VICON).
  The thrust of each rotor is calculated by a quadratic model with the coefficient $c_t$, and rotor speeds (RPM) measured by the electronic speed controller (ESC).
  While the failures are triggered manually for our experiments, a fault detection algorithm (e.g.,~\cite{freddi2010actuator,amoozgar2013experimental}) could replace the mock fault detector.}
  \label{fig:control_diagram}
\end{figure}

For real-world experiments, we validate the proposed algorithm in an instrumented flight arena~\cite{uzh_robotics_and_perception_group_worlds_2021}, which is equipped with a VICON motion tracking system providing the position and orientation of the quadrotor at \SI{400}{\hertz}.
The quadrotor we used in the experiments is shown in Fig. \ref{fig:drone_photo}.
It is actuated by four motors equipped with 5-inch propellers, each providing a maximum of \SI{8.5}{\newton} lift force.
The state estimation and control runs on an NVIDIA Jetson TX2 system-on-a-chip with a frequency of \SI{150}{\hertz}.
The low-level control is done with a Radix FC\footnote{\url{https://www.brainfpv.com/product/radix-fc/}} flight control, which also sends the IMU and motor speed measurements to the Jetson at \SI{500}{\hertz}.
The quadrotor is powered with a 4-cell LiPo battery and has a total mass of \SI{0.75}{\kg}.
Fig.~\ref{fig:control_diagram} presents the system diagram for real-world experiments.

\subsection{Rotor Failure during Hovering Flight}
In the first experiment, we trigger the rotor failure while the quadrotor is hovering.
After the failure, the reference is set to hover at the same position.

\begin{figure}[t]
  \centering
  \includegraphics[width=\columnwidth]{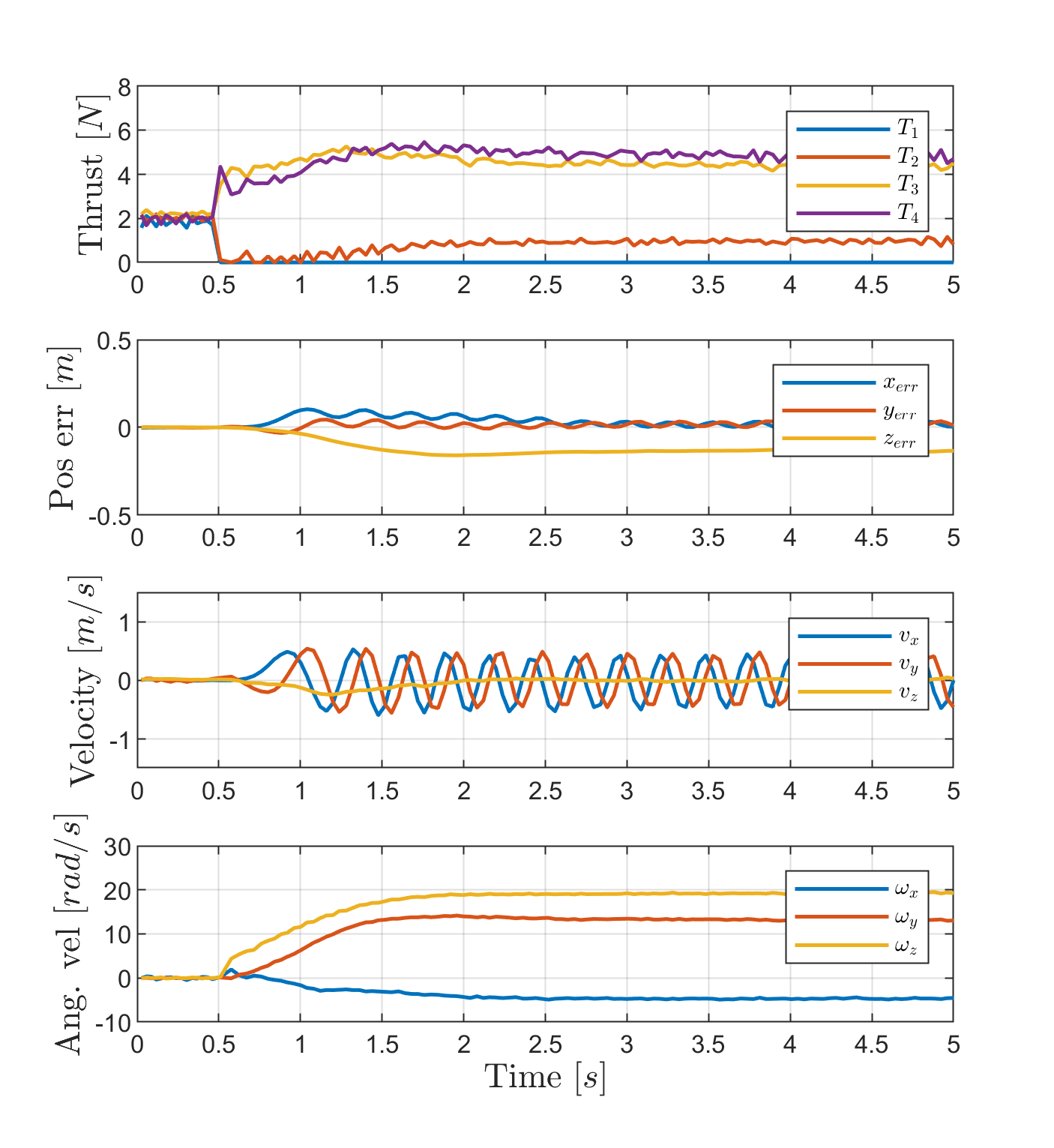}
  \caption{Time history of quadrotor states controlled by the proposed fault-tolerant controller. Failure of the 1st rotor $\left(T_1\right)$ is triggered at $t=\revise{\SI{0.5}{\s}}$. The damaged quadrotor is then stabilized at its original position.}
  \label{fig:hoverfail}
\end{figure}

Fig.~\ref{fig:hoverfail} shows the quadrotor states in this test.
The failure is triggered at $t=\revise{\SI{0.5}{\s}}$, and the quadrotor subsequently starts yaw spinning. Even so, the position in $x$ and $y$ directions are effectively controlled.
Within \SI{2}{\second} after the failure, the quadrotor enters the stable rotating equilibrium.
We observe that the gain selection of the NMPC controller can influence the behavior in the controlled flights. In general, with higher weights on the attitude error and pitch and roll rates, the controller leads to a dynamic equilibrium with a smaller inclination angle and a higher yaw rate. 

\subsection{Trajectory Tracking under Rotor Failure}
Instead of hovering at a single waypoint, we test the performance of our controller in tracking agile trajectories with the top speed at \revise{\SI{5}{\meter\per\second}}. 
A "lemniscate" reference trajectory is selected, which is defined by the following parametric equation:
\begin{equation}
\begin{aligned}
  x(t) &= 4 \sin(\omega t), &
  y(t) &= 2 \sin(2\omega t),
\end{aligned}
\end{equation}
where $\omega$ is a constant to control the reference velocity.
We tested with $\omega_1 = 0.3535$ and $\omega_2 = 0.8838$, such that the maximum velocity of the reference trajectory is \SI{2}{\meter\per\second} and \SI{5}{\meter\per\second} respectively.

\begin{figure}[t]
  \centering
  \includegraphics[width=0.8\columnwidth]{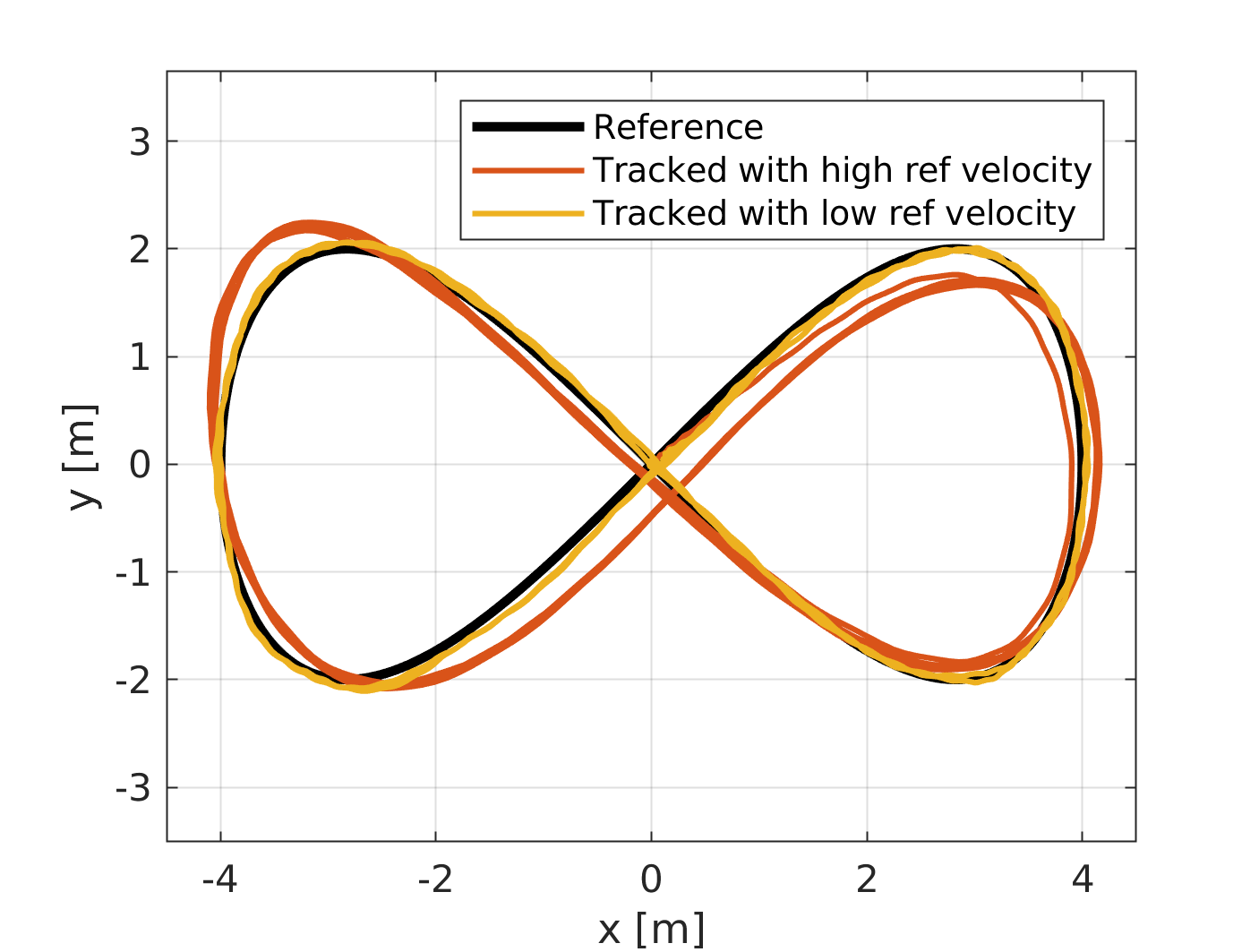}
  \caption{Reference trajectory and tracked trajectory with maximum reference velocity \SI{5.0}{\meter\per\second} and \SI{2.0}{\meter\per\second} respectively}
  \label{fig:lemniscate_trajectory}
\end{figure}

The tracking performance of these two trajectories is shown in Fig.~\ref{fig:lemniscate_trajectory}.
The quadrotor is able to track the slower trajectory accurately, despite the complete loss of a single rotor. As is expected, tracking the faster trajectory leads to a higher tracking error. Be that as it may, this experiment validates the performance of tracking a pre-designed trajectory at up to 5~m/s, showing the ability of a quadrotor to navigate home or finish the delivery task even if a rotor has failed.

\begin{figure}[t]
  \centering
  \includegraphics[width=\columnwidth,trim={0 5 0 0},clip]{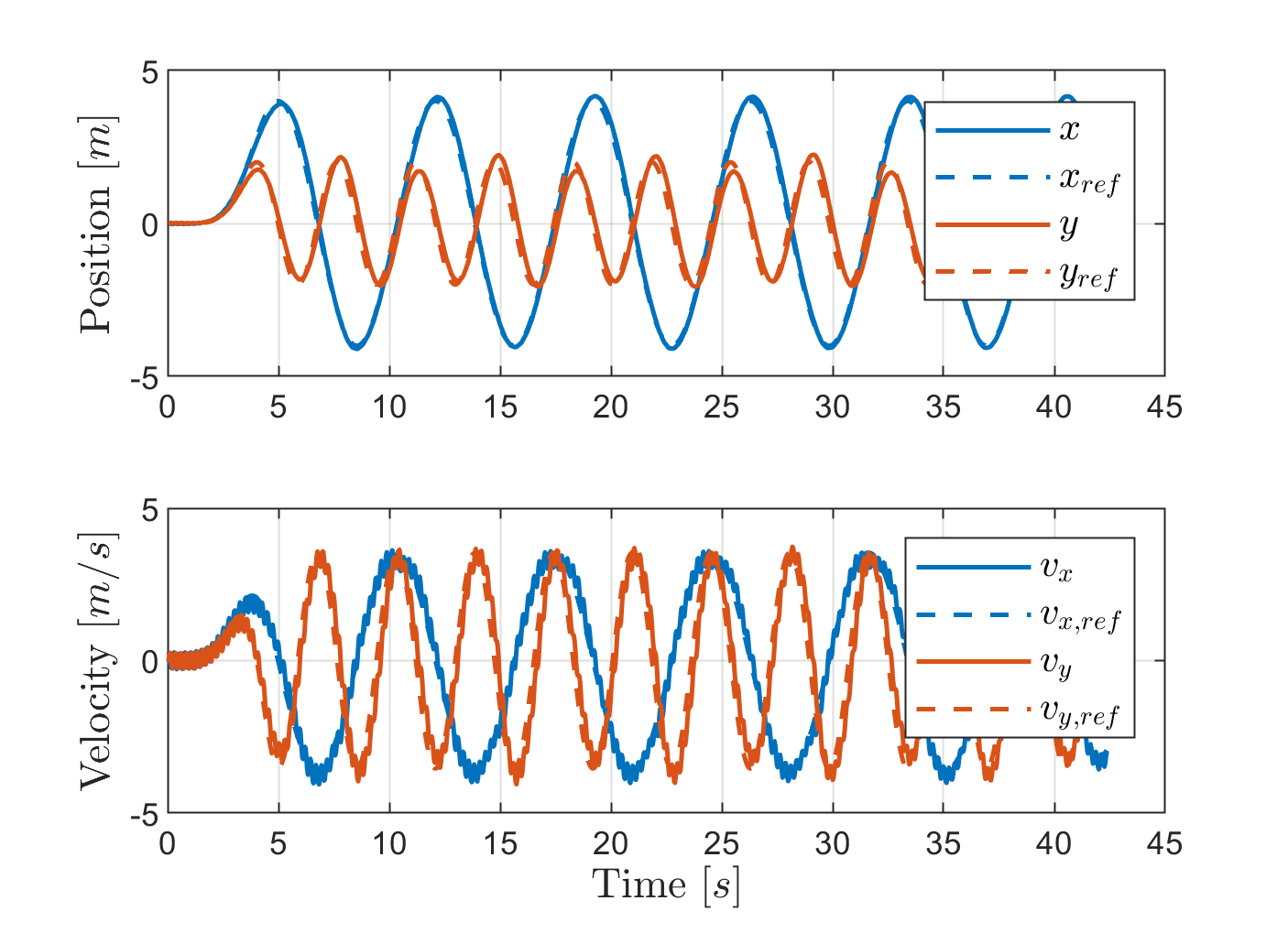}
  \caption{Quadrotor position and velocity during tracking the trajectory with a max velocity at \SI{5}{\meter\per\second}}
  \label{fig:lemniscate_posvel}
\end{figure}
Fig. \ref{fig:lemniscate_posvel} shows the position and velocity of the center of gravity.
According to Fig. \ref{fig:lemniscate_posvel}, the desired position of the quadrotor is followed with a small delay.
The high-frequency oscillations in the velocity plots are caused by the spinning motion of the quadrotor.
We hypothesize that the inadequate tracking performance of the quadrotor is caused by the model uncertainties from aerodynamic effects~\cite{sun_aerodynamic_2019}. 
Even though an INDI controller is implemented, it does not compensate for the aerodynamic forces on the damaged quadrotor.
The effect of these aerodynamics on the trajectory tracking performance still needs to be further justified.


\subsection{Rotor Failure during Agile Flights}
As mentioned before, a major advantage of the NMPC is the ability to exploit the full nonlinear dynamics of the quadrotor. Thus it can recover the damaged quadrotor from extreme conditions largely deviated from the hovering condition. Similarly to the simulations performed in Sec.~\ref{sec:simulations}, we conducted two real-world experiments:
\begin{itemize}
    \item i) The rotor failure occurs when the drone is flying upside-down on the apex of a vertical flipping maneuver.
    \item ii) The rotor failure occurs when the drone is tracking an agile racing trajectory.
\end{itemize}

\begin{figure}[t]
  \centering
  \includegraphics[width=\columnwidth]{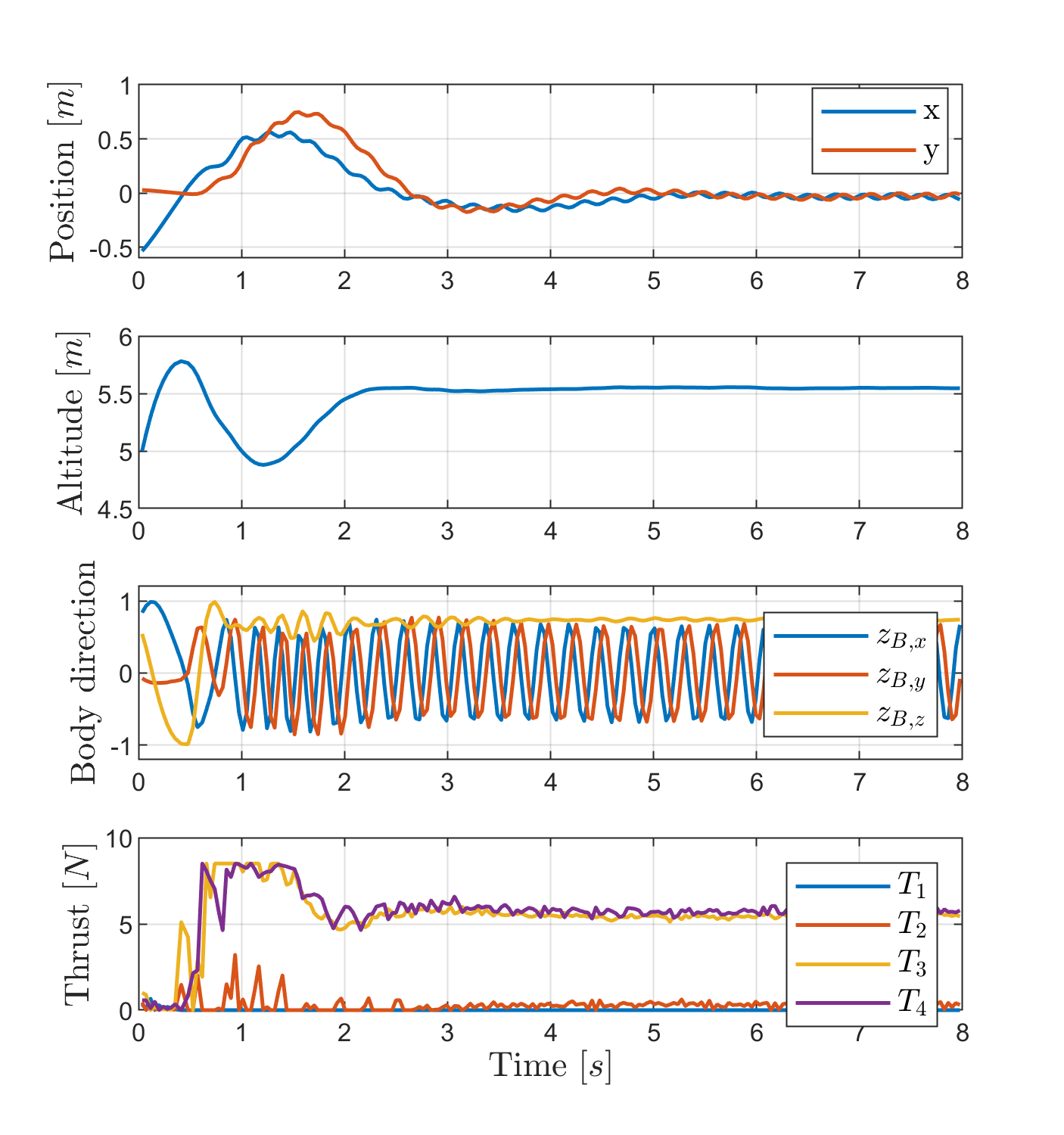}
  \caption{Time history of quadrotor states during the post-failure recovery process from an upside-down condition. Failure is triggered at $t=\revise{\SI{0.5}{\second}}$. Positional origin is shifted in $x$ and $y$ directions to the point where the rotor failure happened.}
  \label{fig:fliprecovery}
\end{figure}

Fig.~\ref{fig:fliprecovery} shows the quadrotor trajectory in the first scenario, namely recovery during a flipping maneuver.
A snapshot of the experiment is shown in Fig.~\ref{fig:fliprecovery_snapshot}.
The first two plots in Fig.~\ref{fig:fliprecovery} present the position of the quadrotor during the post-failure recovery process; the third plot shows the quadrotor orientation with curves showing the $x$, $y$, and $z$ components of the thrust direction $\bm{z}_B$ in the inertial frame; and the last plot shows the thrust command of the controller.
In this experiment, the quadrotor reaches the upside-down orientation at $t\approx \SI{0.5}{\second}$. At this moment, failure of the 1st rotor is triggered, and the quadrotor is immediately commanded to recover to the hovering state.
Right after the moment when the failure happens, $T_3$ is set as 5~N while $T_2$ and $T_4$ are commanded as zero, such that the quadrotor can quickly flip to recover the orientation.
Then the controller increases $T_4$ to stop the drone from flipping. Finally, the high thrusts on both $T_3$ and $T_4$ lead to a high yaw rate that further stabilizes the quadrotor leveraging the gyroscopic effect, and the position is consequently controlled.
The above process only lasts for \SI{2}{\second}, and the height of the damaged quadrotor is only dropped by \SI{0.9}{\meter}.

\begin{figure}[t]
    \centering
    \includegraphics[width=\columnwidth]{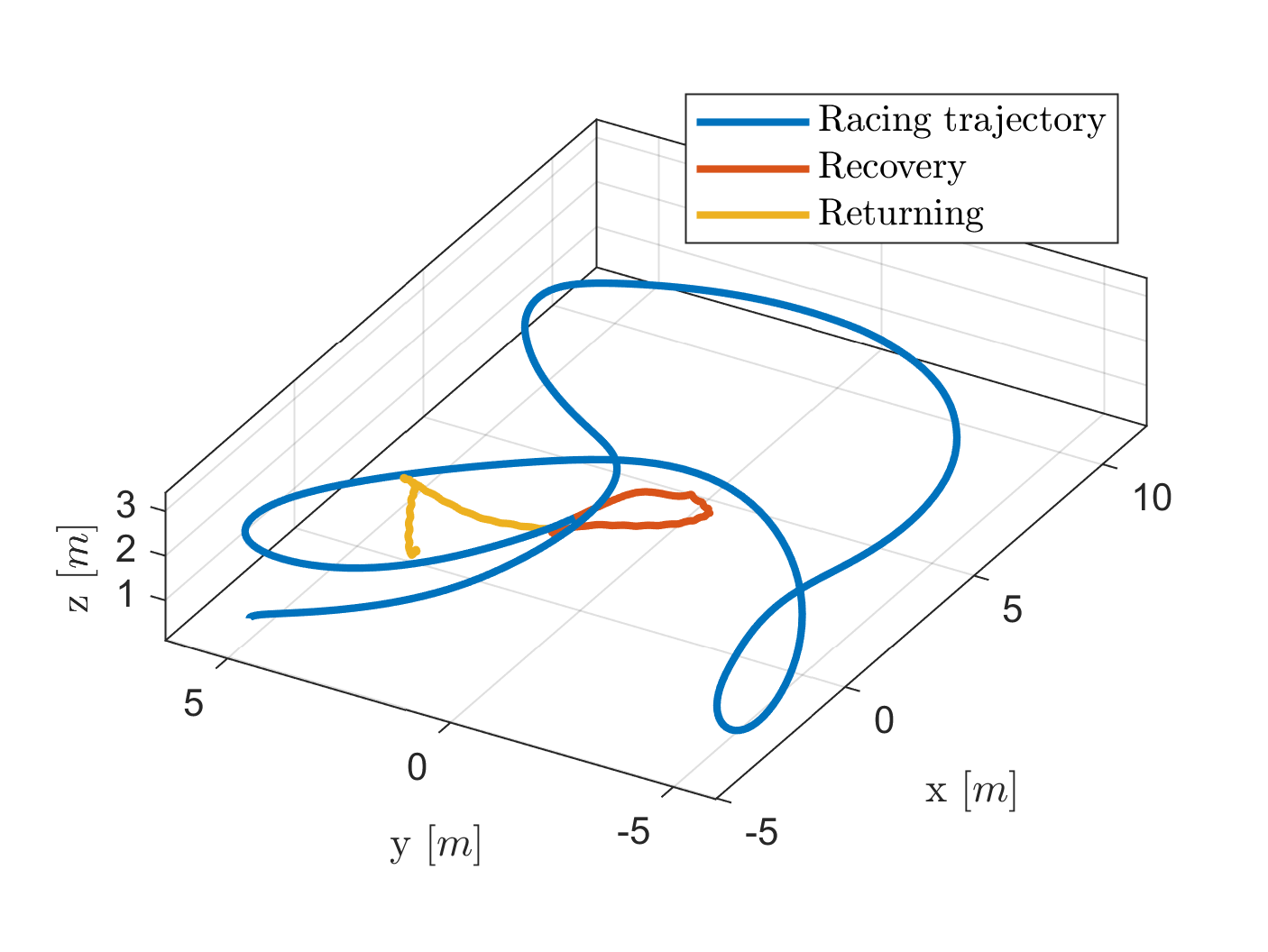}
    \caption{The 3D trajectory of the quadrotor recovering when experiencing a rotor failure in a racing trajectory}
    \label{fig:cpcfail_traj}
\end{figure}

Finally, we tested triggering the rotor failure while tracking an agile racing trajectory at speed up to \revise{\SI{16}{\meter \per \second}}.
Fig. \ref{fig:cpcfail_traj} shows the trajectory of the quadrotor during the experiment.
Failure of the 1st rotor is triggered when the quadrotor reaches a designated position. At this moment, the quadrotor is almost \revise{\SI{90}{\degree}} inclined, and is flying at over \revise{\SI{7.5}{\meter\per\second}}.
Then the fault-tolerant controller successfully recovers the pose of the drone in spite of the large translational speed and the resultant aerodynamic effects. The entire recovery procedure lasts for \SI{3}{\second}.
In the end, the damaged quadrotor is commanded to return and land.

\begin{figure}[t]
    \centering
    \includegraphics[width=\columnwidth]{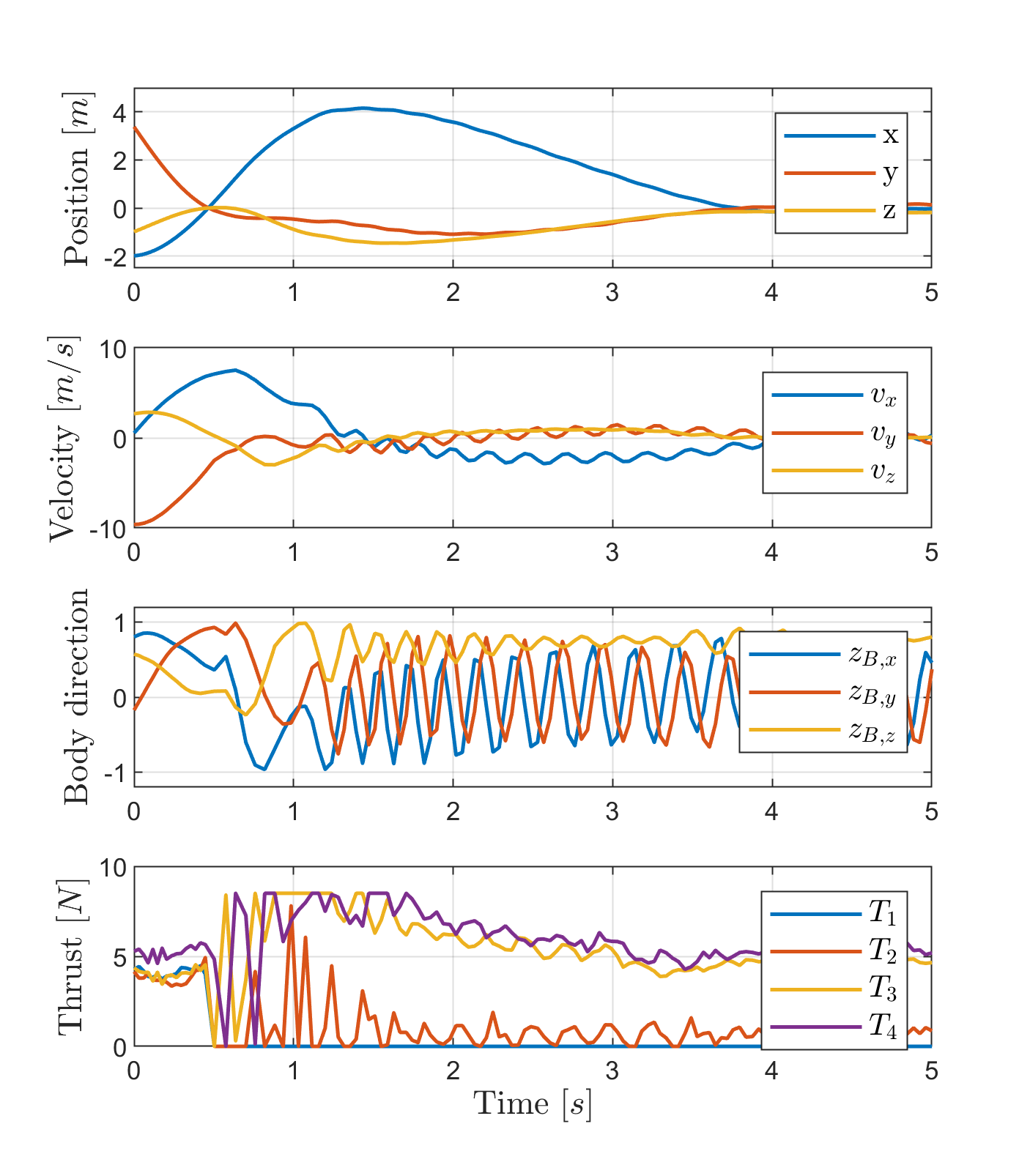}
    \caption{Time history of the quadrotor states during the recovery process  when experiencing a rotor failure in a racing trajectory. Failure is triggered at $t=\revise{\SI{0.5}{\second}}$. Positional origin is shifted to the point where the rotor failure happened.}
    \label{fig:cpcfail_state}
\end{figure}

\section{Conclusions}\label{sec:conclusions}
In this work, we proposed a nonlinear model predictive control framework for quadrotor fault-tolerant flight.
The NMPC method can effectively stabilize the quadrotor subjected to the complete failure of a single rotor.
The proposed method does not use a cascaded structure, nor rely on linear assumptions.
Instead, it takes the full nonlinear model of the quadrotor, and takes into account the constraints of each rotor.
\revise{Extensive tests against traditional nonlinear control method have showed that the NMPC is more robust to the initial condition where the rotor failure happens. The proposed method can recover the damaged quadrotor when a single rotor failure happens at arbitrary orientation or during an agile flight. The method is also able to control the position of the damaged quadrotor and even follow a trajectory.}



\section*{ACKNOWLEDGMENT}
The authors would like to thank Thomas Längle, Leonard Bauersfeld, and Jiaxu Xing for their help in experiments,  photo rendering, and data collection.

\bibliographystyle{ieeetr}
\bibliography{IEEEabrv, RotorfailMpc}

\end{document}